%% file: acl_latex.tex
\pdfoutput=1

\documentclass[11pt]{article}

\usepackage[]{acl}

\usepackage{times}
\usepackage{latexsym}
\usepackage{graphicx}

\usepackage[ruled,vlined]{algorithm2e} 
\usepackage{algorithmic}

\usepackage{multirow}
\usepackage{amssymb} 
\usepackage{booktabs} 

\usepackage{arydshln}   

\usepackage[T1]{fontenc}

\usepackage[utf8]{inputenc}

\usepackage{microtype}
\usepackage{xspace}

\newcommand{\OURS}{OlympiadBench\xspace}

\usepackage{inconsolata}

\title{\OURS: A Challenging Benchmark for Promoting AGI with \\ Olympiad-Level Bilingual Multimodal Scientific Problems}

\author{Chaoqun He$^{1}$, Renjie Luo$^{2}$, Yuzhuo Bai$^{1}$, Shengding Hu$^{1}$, Zhen Leng Thai$^{1}$\\
\textbf{Junhao Shen$^{1}$, Jinyi Hu$^{1}$, Xu Han$^{1*}$, Yujie Huang$^{1}$, Yuxiang Zhang$^{3}$}\\
\textbf{Jie Liu$^{3}$, Lei Qi$^{3}$, Zhiyuan Liu$^{1}$\thanks{Corresponding authors: Xu Han and Zhiyuan Liu.}}\textbf{, Maosong Sun$^{1}$}\\
\textsuperscript{1}{\small Dept. of Comp. Sci. \& Tech., Institute for AI, Tsinghua University, Beijing, China}\\
\textsuperscript{2}{\small Institute of Artificial Intelligence, Beihang University, China}\\
\textsuperscript{3}{\small Wisdom Way AI Lab, China}\\
{\tt\small \{hcq21,byz22\}@mails.tsinghua.edu.cn, renjie.luo@outlook.com, \{hanxu2022,liuzy\}@tsinghua.edu.cn}
}

\begin{document}
\maketitle
\begin{abstract}
Recent advancements have seen Large Language Models (LLMs) and Large Multimodal Models (LMMs) surpassing general human capabilities in various tasks, approaching the proficiency level of human experts across multiple domains. 
With traditional benchmarks becoming less challenging for these models, new rigorous challenges are essential to gauge their advanced abilities. 
In this work, we present OlympiadBench, an Olympiad-level bilingual multimodal scientific benchmark, featuring 8,476 problems from Olympiad-level mathematics and physics competitions, including the Chinese college entrance exam. 
Each problem is detailed with expert-level annotations for step-by-step reasoning. Evaluating top-tier models on OlympiadBench, we implement a comprehensive assessment methodology to accurately evaluate model responses. 
Notably, the best-performing model, GPT-4V, attains an average score of 17.97\% on OlympiadBench, with a mere 10.74\% in physics, highlighting the benchmark rigor and the intricacy of physical reasoning. 
Our analysis orienting GPT-4V points out prevalent issues with hallucinations, knowledge omissions, and logical fallacies. We hope that our challenging benchmark can serve as a valuable resource for helping future AGI research endeavors. The data and evaluation code are available at \url{https://github.com/OpenBMB/OlympiadBench}
\end{abstract}

\section{Introduction}

\begin{figure}[ht]
    \centering
    \includegraphics[width=1\linewidth]{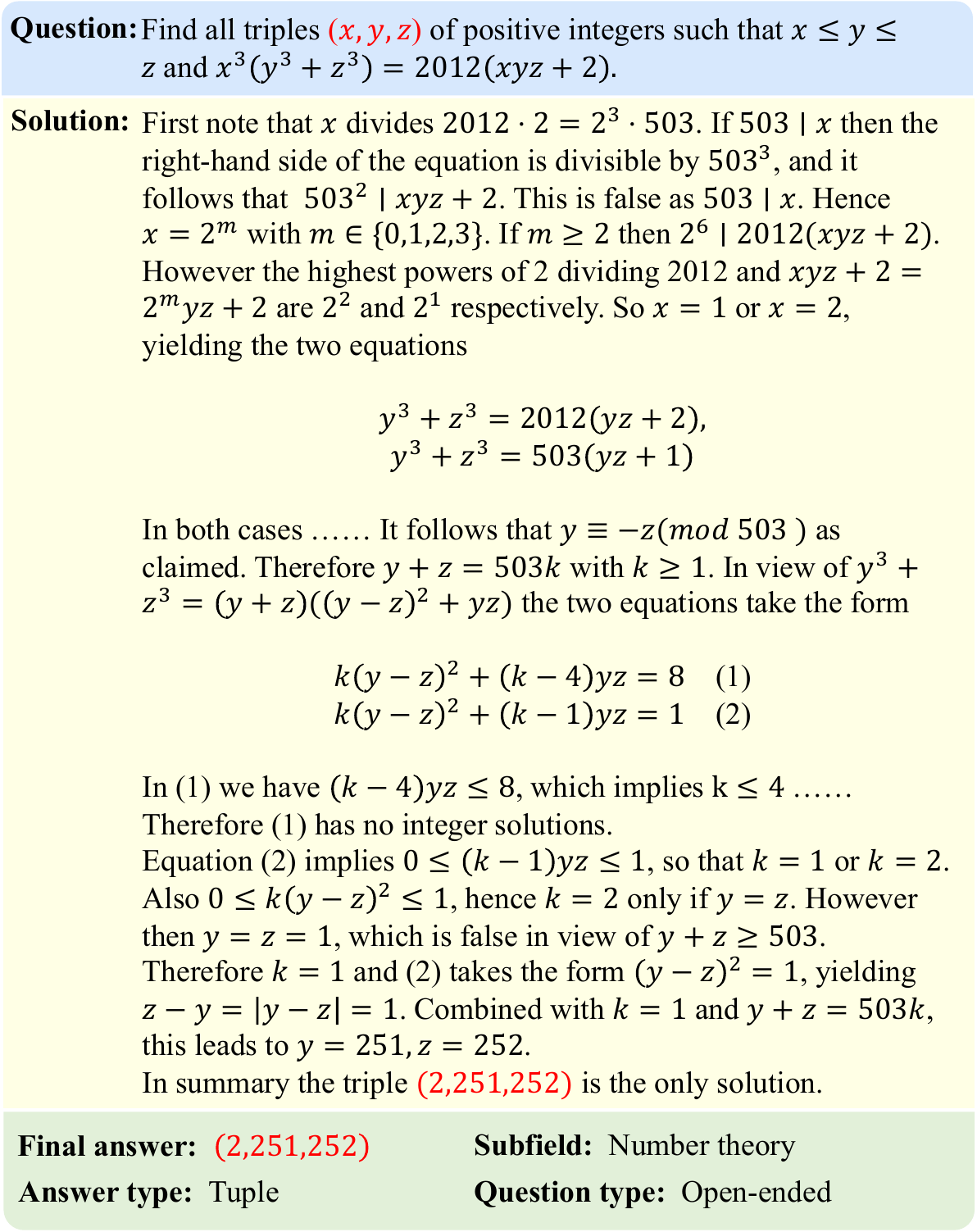}
    \caption{An example of IMO in OlympiadBench. Solving this example requires AI systems to span different mathematical domains and conduct advanced reasoning.}
    \label{fig:fig1}
    \vspace{-1em}
\end{figure}

Large Language Models (LLMs) have demonstrated remarkable capabilities across various tasks such as text generation~\cite{zhao2023survey}, code generation~\cite{zan-etal-2023-large} and mathematical reasoning~\cite{lu2023survey,zhou2023solving}, garnering significant attention from both academia and industry~\cite{wei2022emergent,zhao2023survey,bubeck2023sparks}. 
The most powerful models such as GPT-4~\cite{openai2023gpt4} and Gemini Ultra~\cite{geminiteam2023gemini} have even surpassed oridinary human level on a wide variety of benchmarks such as MMLU~\cite{hendrycks2020measuring}, MMMU~\cite{yue2023mmmu}, and even surpassing human expert in many area.
These results show a promising future that LLMs can serve as proficient assistants for human scientists~\cite{nguyen2023brief,qiu2023transfer}.
Among the array of expert-level skills exhibited by LLMs, scientific reasoning consistently emerges as one of the most brilliant, showcasing some of the most distinguished intellectual properties that experts possess. Therefore, this paper primarily focuses on mathematical and physical reasoning.

In recent years, several benchmarks related to mathematics have been proposed, such as the dataset GSM8K~\cite{cobbe2021gsm8k} as well as the dataset MATH~\cite{hendrycks2021measuring}. However, these benchmarks, are primarily developed before the advent of highly capable LLMs, and now lack sufficient challenge for the latest models. For instance, GPT-4 with prompting techniques\cite{zhou2023solving} has achieved a 97.0\% success rate on GSM8K and 84.3\% on MATH. The rapid evolution of LLMs may soon lead to saturated results on these benchmarks. Concurrently, LLMs are not yet fully equipped to assist mathematicians in solving complex problems~\cite{collins2023evaluating,zhang2023ai}, nor are they capable of performing expert-level mathematical reasoning independently. This discrepancy underscores the need for more challenging datasets to benchmark future advancements of LLMs in this domain.
Similarly, physics presents comparable challenges for AI to those found in mathematics. Nevertheless, existing benchmarks related to physics~\cite{lu2022learn,arora2023llms,wang2024scibench} are characterized by their relatively low difficulty and limited scope. There is also a significant lack of a rigorous and challenging benchmark in physics.

In addition to the issue regarding the benchmark difficulty, it is important to note that these benchmarks predominantly focus on text. 
This presents a significant limitation, as a wide range of scientific reasoning contexts require multimodal reasoning abilities. 
For example, grasping geometry reasoning in mathematics or understanding experiments designs in physics are scenarios where multimodal reasoning capabilities are crucial. 
Notably, various large multimodal models (LMMs) have been developed~\cite{geminiteam2023gemini,liu2023visual} and demonstrate proficiency on a variety of tasks~\cite{lu2022learn,yue2023mmmu,zhang2024cmmmu,lu2024mathvista}, offering the potential for multimodal scientific reasoning. 
Nevertheless, there is still a lack of sufficient benchmarks to prove whether these LMMs are capable of handling scientific problems.
Consequently, a challenging multimodal benchmark is essential for advancing scientific reasoning tasks\cite{zhang2024mmllms,lu2023survey}.

To address the aforementioned inadequacies, we introduce OlympiadBench, an Olympiad-level bilingual multimodal scientific benchmark. This collection comprises 8,476 math and physics problems sourced from International Olympiads, Chinese Olympiads, and the most challenging segments of the Chinese College Entrance Exam (GaoKao). We download PDF data from official websites and utilize Mathpix\footnote{\url{https://mathpix.com/}} for OCR parsing. 
We meticulously inspect, clean, and revise the data, and further adopt LLMs for deduplication. 
Finally, we annotate the data with crucial information such as answer types and subfields, yielding a dataset that is clean, accurate, and detailed.
As shown in Figure~\ref{fig:fig1}, \OURS features numerous distinct characteristics such as difficulty, free-form generation, expert-level solution annotation, detailed labeling of difficulty, wide-coverage of modality and language, etc. These features are summarized more clearly from Table~\ref{tab:benchmark_comparison}.

We conduct an evaluation of current state-of-the-art LLMs and LMMs on the OlympiadBench. 
The best-performing model, GPT-4V, is a multimodal version of GPT-4 developed by OpenAI that can understand images. Despite its advanced capabilities, GPT-4V achieves a score of only 17.97\% on OlympiadBench, with individual scores of 21.70\% in mathematics and 10.74\% in physics.

Importantly, the experiment results show that LMMs still struggle in computational error, incorrect reasoning or induction.
For the process involved in the correct responses, the process occasionally includes hallucinated reasoning, or choosing a more complex solution when a simpler solution exists.
All these results highlight the substantial challenge OlympiadBench presents to contemporary large models and point the direction of future efforts.

\OURS is inspired by the significant advances made by DeepMind AlphaGeometry~\cite{trinh2024solving}, which nearly matches the proficiency of International Mathematical Olympiad (IMO) gold medalists in geometry proofs. It is clear that \OURS, along with other challenging datasets like the AI-MO challenge\footnote{\url{https://aimoprize.com/}}, will witness and benchmark the swift progress towards expert-level AI assistants for solving scientific problems.

\section{Related Work}

This section gives an overview of the existing datasets in solving mathematics and physics problems as well as multimodal datasets.


\textbf{Mathematics Benchmarks.} Solving mathematics problems and proving theorems in natural languages has been a key research focus in machine learning and natural language processing since the 1960s~\cite{bobrow1964natural}. Previous benchmarks~\cite{koncel-kedziorski-etal-2016-mawps,wang-etal-2017-deep,ling-etal-2017-program,amini-etal-2019-mathqa,cobbe2021gsm8k,wei2023cmath} focus predominantly on math word problems (WMPs) which involve four basic arithmetic operations with single or multiple operation steps~\cite{lu2023survey}.
Typically, the GSM8K~\cite{cobbe2021gsm8k} dataset targets elementary-level questions within 8 steps of basic arithmetic operations. 
However, these problems are typically text-only~\cite{lu2023survey} and of lower difficulty, with reasoning limited to a few computations. 
As the complexity of the problems rises, some works~\cite{hendrycks2021measuring,frieder2023mathematical,arora2023llms} introduce competition-level problems integrating mathematical logic and background knowledge. 
Yet, these challenging datasets are increasingly being surmounted~\cite{zhou2023solving}. 
Theorem proving is a problem to demonstrate the truth of a mathematical claim (a theorem) through a sequence of logical arguments (a proof)~\cite{lu2023survey}. 
Earlier efforts mainly focused on translating natural language proofs into formal representations, facing significant expertise and labor challenges~\cite{zheng2022minif2f,welleck2021naturalproofs}. 
The emergence of LLMs has facilitated notable advancements in the domain of natural language proof~\cite{jiang2023draft}. \OURS presents mathematical reasoning and theoretical proofs all in natural language with detailed solution annotations.

\begin{table*}[ht!]
\centering
\setlength{\tabcolsep}{3pt}
\renewcommand{\arraystretch}{1.2}
\resizebox{0.9\linewidth}{!}{
\begin{tabular}{lcccccccccc}
\toprule
\multirow{2}{*}{\textbf{Benchmark}} & \multicolumn{2}{c}{\textbf{Subject}} & \multicolumn{1}{c}{\textbf{Multi-}} & \multicolumn{1}{c}{\textbf{Detailed}} & \multicolumn{1}{c}{\textbf{Difficulty}}  & \multicolumn{2}{c}{\textbf{Size}} &  \multicolumn{1}{c}{\textbf{Answer}}  & \multirow{1}{*}{\textbf{Language}} & \multicolumn{1}{c}{\textbf{Question}} \\
\cline{2-3} \cline{7-8}
        & \textbf{Maths} & \textbf{Physics} & \textbf{modal} & \textbf{solution} & \textbf{level} & \textbf{Maths} & \textbf{Physics}  & \textbf{type}  &  \textbf{type} & \textbf{type} \\
\midrule
SciBench   & \checkmark & \checkmark & \checkmark & \checkmark & COL & 217 & 295  & Num  & EN & OE \\
MMMU       & \checkmark & \checkmark & \checkmark & \checkmark & COL & 540 & 443 & Num & EN & MC,OE \\
MathVista  & \checkmark &  & \checkmark &  & - & 1,000 &   & Num & EN & MC,OE \\
ScienceQA  &  & \checkmark & \checkmark &  & H &  & 617 &  & EN & MC \\
SciEval    &  & \checkmark &  &  & - &  & 1,657 & Num & EN & MC,FB,J \\
JEEBench   & \checkmark & \checkmark &  & \checkmark & CEE & 236 & 123 & Num &  EN & MC,OE \\
MMLU       & \checkmark & \checkmark &  &  & COL & 948 & 548 &  & EN & MC \\
AGIEval    & \checkmark & \checkmark &  &  & CEE & 953 & 200 & Num & EN,ZH & MC,FB,OE \\
GSM8K      & \checkmark &  &  & \checkmark & E & 1,319 &  & Num & EN & OE \\
MATH       & \checkmark &  &  & \checkmark & COMP & 5,000 &  & Num,Exp,Tup & EN & OE \\
\hline
\textbf{OlympiadBench} & \checkmark & \checkmark & \checkmark & \checkmark & COMP & 6,142 & 2,334 & ALL & EN,ZH & OE \\

\bottomrule
\end{tabular}
}
\caption{For \textbf{difficulty level}, COMP: Competition, COL: College, CEE: College Entrance Examination, H: High School, E: Elementary School, and we picked the highest level; For \textbf{answer type}, Num: Numeric value, Exp: Expression, Equ: Equation, Int: Interval, Tup: Tuple; For \textbf{language type}, EN: English, ZH: Chinese; For \textbf{question type}, OE: Open-ended, MC: Multiple-choice, FB: Fill-in-the-blank, J: Judgement. For the statistical analysis of quantity and relevant metrics in AGIEval, we exclude 1,000 questions from the MATH benchmark to facilitate a more accurate comparison. The “-” indicates that it cannot be confirmed. Upon comparison, \OURS leads in all aspects.}
\label{tab:benchmark_comparison}
\end{table*}

\textbf{Physics Benchmarks.} Physics questions in SciQ~\cite{welbl-etal-2017-crowdsourcing}, ScienceQA~\cite{lu2022learn} and E-EVAL~\cite{hou2024eeval} are mainly elementary and high school level multiple-choice questions, lacking complex reasoning and computational tasks. In MMLU-STEM~\cite{hendrycks2020measuring} and C-Eval-STEM~\cite{huang2023ceval}, physics questions also adopt a multiple-choice format. JEEBench~\cite{arora2023llms} extends this format to include multistep reasoning with physics knowledge, yet it is limited in scope and purely text-only. 
TheoremQA~\cite{chen-etal-2023-theoremqa} is the first theorem-driven question-answering dataset. Curated by domain experts, it contains 800 high-quality questions that encompass 350 theorems from Mathematics, Physics, Electrical Engineering and Computer Science (EE\&CS), and Finance.
SciEval~\cite{sun2023scieval} consists of a total of about
18,000 challenging scientific questions, spanning three important basic science fields: chemistry, physics and biology. SciBench~\cite{wang2024scibench} and OCWCourses~\cite{lewkowycz2022solving} offer college-level physics questions in free-response formats, where SciBench contains multimodal information. In contrast, \OURS escalates in difficulty, diversifies in question types, and surpasses in volume, setting a new benchmark for complexity and variety in the domain.

\textbf{Multimodal Benchmarks.} For assessing multimodal capability, works such as Geometry3K~\cite{lu-etal-2021-inter}, GeoQA~\cite{chen-etal-2021-geoqa}, GeoQA+~\cite{cao-xiao-2022-augmented}, and UniGeo~\cite{chen-etal-2022-unigeo} have employed multimodal information for tackling geometric problems, integrating natural language descriptions with diagrams. 
ScienceQA~\cite{lu2022learn}, MMMU~\cite{yue2023mmmu}, CMMMU~\cite{zhang2024cmmmu} and CMMU~\cite{he2024cmmu} are multimodal, multi-discipline evaluation sets, encompassing a broad range of subjects. 
MathVista~\cite{lu2024mathvista} integrates 28 existing and 3 newly constructed multimodal datasets involving mathematics, aiming to establish a benchmark that encapsulates challenges from a variety of mathematical and visual tasks. However, it does not concentrate on delving into the complexity of mathematics problems. 

In summary, we introduce a new benchmark to address these gaps. Table \ref{tab:benchmark_comparison} presents a comparison between \OURS and several related benchmarks, highlighting the significant advantages of \OURS across all aspects.

\section{The OlympiadBench Dataset}

To evaluate the reasoning abilities of LLMs and LMMs in mathematics and physics problems, we have created \OURS, a bilingual and multimodal scientific benchmark at the competition level. This section provides a detailed account of the construction process of \OURS. Summarized statistics
of the dataset is shown in Table \ref{tab:benchmark_statistics}, and more detailed statistics are in Appendix \ref{subsec: statistics of data}.

\begin{table}[ht!]
\centering
\resizebox{0.9\columnwidth}{!}{ 
\begin{tabular}{l r}
\toprule
\textbf{Statistics} & \textbf{Number} \\
\hline
Total Problems & 8,476 \\
\hspace{3mm}* Problems with images & 4,869 (57\%) \\
\hspace{3mm}* Problems with solutions & 8,476 (100\%) \\
Difficulties (CEE: COMP) & 66\%: 34\% \\
EN: ZH & 2,125: 6,351 \\
\hline
Open-ended Questions & 6,728 (79\%) \\
Theorem Proving & 1,748 (21\%) \\
\hline
Math: Physics & 6,142: 2,334 \\
\hspace{3mm}* Maths with images & 2,911\\
\hspace{3mm}* Physics with images & 1,958\\
\hline
Average question tokens & 253 \\
Max question tokens & 3,745 \\
Average solution tokens & 347 \\
Max solution tokens & 4,223 \\
\bottomrule
\end{tabular}
}
\caption{Statistics of \OURS. When calculating tokens, images are not included.}
\label{tab:benchmark_statistics}
\end{table}

\subsection{Design Principle}
\label{sec:design_principle}
The motivation behind \OURS is to establish a benchmark that represents the pinnacle of human intellectual achievement, thereby encouraging researchers of large models to push the boundaries of mathematical and physical reasoning capabilities. We focus on curating challenges that epitomize the highest level of competition worldwide. Specifically, \OURS includes:

\begin{enumerate}
    \item \textbf{Inclusion of Olympiad-Level Problems.} 
    We collect mathematics and physics problems from the International Olympiad competitions, which cater to the most outstanding high school students in a region. These problems are open-ended, differing from traditional multiple-choice or fill-in-the-blank formats. This selection aims to more accurately reflect the complexity of advanced scientific reasoning, providing insight into the actual reasoning process of the models.
    \item \textbf{Provision of Detailed Solutions.} Given the advanced difficulty of these problems, which may exceed the comprehension of individuals without a specialized background in mathematics, each problem is accompanied by expertly crafted solutions that detail the reasoning steps involved. This approach can not only reduces the difficulty of annotation and evaluation but also enhances the accuracy of the solutions provided. 
    Furthermore, detailed expert-level solutions are valuable for research in model reasoning.
    \item \textbf{Incorporation of Visuals.} Recognizing the crucial role of visual information in conveying complex ideas, \OURS incorporates problems that require understanding images, identifying spatial relationships, and other advanced reasoning tasks. This inclusion aims to assess the model's capabilities in interpreting visual data as part of its reasoning process.
    \item \textbf{Minimization of Data Leakage Risks.} To minimize the risk of data leakage, we have sourced problems from official Olympiad competitions, converting them from their original PDF files provided by official websites to the markdown format required. This strategy is aimed at reducing the likelihood of the data being inadvertently incorporated into the pre-training corpora of models.

\end{enumerate}
Through these carefully designed criteria, \OURS aims to not only challenge but also significantly advance the capabilities of models in mathematical and physical reasoning.


\subsection{Data Processing}

The data processing pipeline is structured into three distinct phases: data collection, format conversion \& deduplication, and classification labeling.

\textbf{Data Collection.} \OURS is meticulously compiled from three primary sources: Global Mathematics and Physics Olympiad Problems, Regional and National Chinese Math Competitions, and Gaokao Mock Questions for Mathematics and Physics
~\footnote{Clean and correct college entrance exams as well as mock questions from Wisdom Way AI Lab.}.
Each chosen for its distinct advantages in creating a robust and comprehensive benchmark for evaluating LLMs and LMMs in mathematical and scientific reasoning. Their challenges progressively increase in difficulty, not only distinguishing the reasoning capabilities of models of various sizes but also offering guidance on scaling laws~\cite{kaplan2020scaling} for specialized models in these domains.

\textbf{Format Conversion and Deduplication.} After collecting all PDF files, we utilize the Mathpix tool for OCR recognition and convert them into markdown format. However, no conversion process is flawless, necessitating manual verification by our team members between the original PDF files and the converted Markdown texts. The Markdown texts are further structured into a format akin to "Problem—Solution—Answer", employing its markup language for text organization. Subsequently, we leverage a specialized small-scale language model~\footnote{\url{https://huggingface.co/Laurie/Bloom1b7-deepspeed-chat-Chinese-math}} trained on mathematical symbol corpora for vectorizing the data and performing deduplication based on cosine similarity.

\begin{figure}[htbp]
    \centering
    \includegraphics[width=0.7\linewidth]{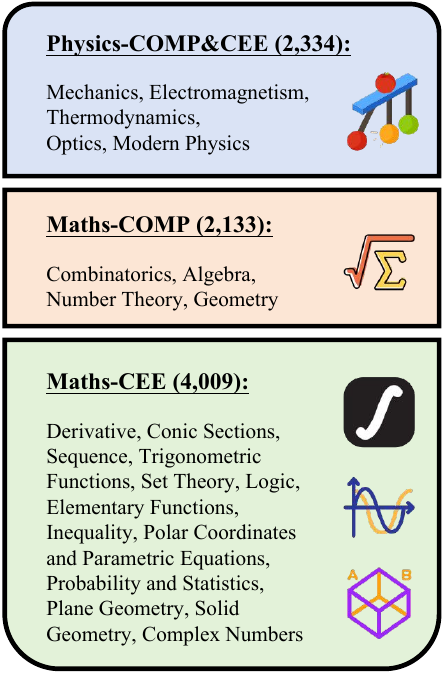}
    \caption{Subfields Distribution of OlympiadBench}
    \label{fig:subfields of OympiadBench}
\end{figure}

\textbf{Classification Labeling.} We note that both mathematics and physics problems predominantly comprise two types of questions: the open-ended problems and the theorem proving problems. We also note that the dataset, enriched by both Olympiad and national examination questions, covers a broad spectrum of subfields, as illustrated in Figure \ref{fig:subfields of OympiadBench}. Therefore, we manually annotate each question with topic and problem type annotations.







\begin{table}[ht!]
\centering
\resizebox{0.8\columnwidth}{!}{
\begin{tabular}{lc}
\toprule
\textbf{Answer type}     & \textbf{Example}                         \\ \hline
Numeric &      1/4                             \\
Expression      & \(x=(1/2)at^2\)                \\
Equation        & \(x^2+y^2=1\)                   \\
Tuple        & \((x,y,z)=(0,0,0)\) \\
Interval        & \((-\infty,-1)\cup(1,+\infty)\)             \\ 
\bottomrule
\end{tabular}
 }
\caption{Examples of the five answer types}
\label{examples of answer types}
\end{table}

\subsection{Data Characteristics}
In contrast to previous benchmarks, OlympiadBench unveils two unique characteristics within its dataset: the incorporation of Progressive Problems in Physics and the categorization of answers to most open-ended questions into a limited number of types.

\textbf{Progressive Problems in Physics.} In physics competitions such as the International Physics Olympiad (IPhO), problems are often structured around a common material or scenario, with subsequent questions potentially relying on the answers or information from previous questions. One example is given in Figure \ref{fig:ipho2021-problem2-part1} and Figure \ref{fig:ipho2021-problem2-part2}. This design characteristic is commonly referred to as "progressive problems". By linking a series of questions together, progressive problems require participants to apply their knowledge and skills comprehensively to gradually solve more complex issues. This type of question design aims to test students' depth of understanding, application capabilities, and innovative thinking, rather than just basic knowledge. To better utilize this feature, we compile the material, questions, and their answers into the 'context' field for each set of progressive problems.

\textbf{Answer Type Classification.} Whether in mathematics or physics, the answers to problem requiring definitive response can largely be categorized into the following types: numeric, expression, equation, interval, and tuple. Simple examples of these can be seen in Table \ref{examples of answer types}.

\subsection{Automatic Scoring Pipeline}
\label{subsec:pipeline}
We design an automated scoring pipeline (see Algorithm \ref{pipeline algorithm}) to evaluate model-generated answers across complex fields like mathematics and physics, where answers vary from numbers to equations. This method simplifies answers into two categories: numeric values, handled through floating-point operations, and symbolic expressions, requiring symbolic computation.

For equations, we ensure all terms are on one side before dividing to check for mathematical equivalence. Intervals and tuples are compared by extracting and evaluating each element. Numeric answers are verified against a small tolerance of error, defaulting to 1e-8 but adjustable for physics problems to allow for a specific error margin. For expressions, we use the SymPy\footnote{\url{https://www.sympy.org}} library to confirm if the subtraction of two expressions approaches zero, indicating correctness.

\input{experiment}

\input{analysis}


\section{Discussion and Future Work}

Here, we discuss the automated evaluation of theorem problems and the future research directions in advanced disciplines.

\textbf{Evaluation of Theorem's Proof Problems.} The inability to automatically evaluate theorem's proof problems remains a significant challenge today. Currently, mainstream methods for the automatic evaluation of proofs require formalization, necessitating domain expertise and background knowledge. Exploring how to automatically evaluate natural language proofs represents an important research direction. Our dataset includes expert-level comprehensive annotations in a fusion of natural language and LaTeX formats, making it a high-quality resource for research in natural language proof problems and fostering further development. 

\textbf{Expansion of Disciplines.} Mathematics, much like physics, serves as a litmus test for artificial intelligence, requiring a strong foundational knowledge, rigorous high-level computations, and precise logical reasoning. Currently, large models still face significant challenges in mathematics and physics, which are critical hurdles that must be overcome for the development of AGI. 
In our future work, we will integrate additional fields such as geography, biology, and chemistry to provide a more innovative and comprehensive evaluation of a model's reasoning capabilities.

\section{Conclusion}

We create \OURS, an Olympiad-level bilingual multimodal scientific benchmark to assess the capabilities of large models in mathematics and physics reasoning. Each problem is detailed with expert-level annotations for step-by-step reasoning. In our benchmarking, we provide a detailed analysis of model performance, pinpointing prevalent error types and potential areas for enhancement. This significant and challenging effort fills a notable void, and we intend to open-source the benchmark to advance AGI and scientific reasoning research. Future efforts will focus on gathering more challenging questions and broadening the range of subjects to further develop rigorous scientific benchmarks.


\section*{Acknowledgements}
This work is supported by the National Key R\&D Program of China (No.2022ZD0116312), National Natural Science Foundation of China (No. 62236004), Quan Cheng Laboratory (Grant No. QCLZD202301) and Institute Guo Qiang at Tsinghua University.

\section*{Ethical Considerations}

In this paper, we introduce \OURS, a highly challenging bilingual, multimodal scientific benchmark aimed at evaluating the mathematical and physical reasoning of large models now and AGI in the future. The paper outlines the dataset construction, including data gathering, OCR processing, cleansing, deduplication, and detailed annotation. \OURS's data, derived exclusively from official sources, substantially reduces the likelihood of pre-training data leakage. We offer precise annotations for each problem and have implemented an exhaustive evaluation script for more accurate model performance assessment. Additionally, being bilingual and providing expert-level reasoning annotations for every question, \OURS serves as a crucial resource for propelling AGI's prowess in scientific reasoning. Committed to environmental sustainability, we intend to release the dataset and accompanying scripts publicly to cut down on unnecessary carbon footprint. In experiments, we comply with all licenses for models and data.

\section*{Limitations}

In pursuit of understanding the logical reasoning abilities of LLMs and LMMs within the multimodal domains of mathematics and physics, we develop \OURS, a challenging bilingual multimodal scientific benchmark. Despite filling a notable void, this work acknowledges inherent limitations. First, in the \OURS, some questions feature answers that require categorical discussions or textual descriptions, such as proofs, which currently cannot be assessed using regular expressions or tools like SymPy at the code level and necessitate manual review. However, this data holds significant research value. Secondly, the automated scoring system we propose cannot perform specific analysis based on the particulars of each question. It makes logical judgments solely based on the two symbols or numerical expressions inputted, without integrating any special constraints that may exist within the actual problem context. What's more, the development of datasets for multimodal scientific reasoning requires extensive manual effort in gathering and annotating data, which constrains the diversity and difficulty of multimodal scientific challenges. As a result, this hampers AI's capacity to learn from and address more intricate scenarios.

\bibliography{custom}

\newpage

\appendix

\section{Dataset Details}
\label{sec:appendix}


\begin{table*}[ht]
\centering
\begin{tabular}{lccc}
\toprule
\textbf{Subject} & \textbf{Source} & \textbf{Coverage Years} & \textbf{Number} \\ 
\hline
\multirow{5}{*}{Maths}   & IMO    & 2006-2022                                                                   & 509         \\
                         & RMM    & 2011, 2013, 2015-2019, 2021, 2023                                           & 53          \\
                         & ARML   & 2009-2014, 2019, 2023                                                       & 505         \\
                         & EMC    & 1998-2023                                                                   & 364         \\
                         & EGMO   & 2013, 2015-2023                                                             & 64          \\ \hline
\multirow{6}{*}{Physics} & IPhO   & 1984, 1986-1990, 2008-2012, 2014-2016, 2018-2019, 2021 & 381         \\
                         & APhO   & 2013-2015                                                                   & 200         \\
                         & EPhO   & 2019-2022                                                                   & 17          \\
                         & USAPhO & 2017-2021                                                                   & 113         \\
                         & PUPC   & 2020-2022                                                                   & 65          \\ 
                         & OPhO   & 2020-2023                                                                   & 132          \\ 
\bottomrule
\end{tabular}
\caption{Summary of Problems in Math and Physics Competitions, with full acronyms listed in the Table \ref{acronym illustration}}
\label{tab:span year of OlympiadBench}
\end{table*}

\subsection{Data Sources}
Our data sources can be split into the following three parts:
\begin{enumerate}
\item \textbf{Global Mathematics and Physics Olympiad Problems.} The Mathematics and Physics Olympiad problems are globally recognized for their complexity and quality. These problems often require multiple methods of solution and the ability to integrate sub-disciplines from within the broader fields of mathematics and physics. The participants in these competitions represent some of the most proficient individuals worldwide in logical reasoning within mathematics and physics. This not only sets a high standard for problem-solving but also fosters a diverse set of analytical skills that are crucial for the advancement of large models.

\item \textbf{Regional and National Chinese Mathematics Competitions.} In addition to maintaining a high level of difficulty, regional competitions and the CMO introduce elements specific to the Chinese context. This inclusion is instrumental in furthering the development and research of Chinese-oriented and multilingual large models. By encompassing a wide array of mathematics and physics problems, these competitions provide a unique opportunity to develop models that are adaptable and proficient across different mathematical queries, enhancing their versatility and effectiveness.

\item \textbf{Gaokao Mock Questions for Mathematics and Physics.} Given that the resolution of Olympiad-level problems typically necessitates models with substantial parameter sizes, we also incorporate Gaokao simulation problems to evaluate smaller models' capabilities in answering free-form mathematics and physics questions.
\end{enumerate}

The integration of data from Gaokao simulation problems, regional and national competitions, to the global Olympiads constructs a smooth difficulty transition curve. This methodology not only distinguishes the mathematical and physical problem-solving capabilities of different models but also provides guidance on the scaling laws for models specialized in these domains.


\subsection{Data Curation Process}
\label{subsec: statistics of data}
Our initial step involves a comprehensive survey of well-known Olympiad competitions, and the list of which is accessible through the AoPS community platform~\footnote{\url{https://artofproblemsolving.com/community/c13}}. We cataloged these competitions based on several criteria: difficulty level, volume of questions, availability of materials in public domains, language, discipline, and coverage years. Following the design principles outlined in Section \ref{sec:design_principle}, we meticulously select specific contests and years that not only adhere to our dataset design criteria but also try to span the widest possible range of years (Table \ref{tab:span year of OlympiadBench}).

In the format conversion phase, we also manually annotated the subfield of each question in maths or physics, with their distribution detailed in Table \ref{tab:number-subfield}.


\subsection{Example of Progressive Problem in Physics}
\label{subsubsec:progressive problem}
Figures \ref{fig:ipho2021-problem2-part1} and \ref{fig:ipho2021-problem2-part2} present a sequential challenge from the International Physics Olympiad (IPhO) 2021, illustrating the intricacies of progressive problem-solving in a competitive context. This particular problem set exemplifies a common trait in advanced physics competitions: the dependency of many questions on the solutions and materials of preceding ones. These dependencies are sometimes explicit, but most are implicit, weaving a complex web of interconnected knowledge and reasoning.

An explicit instance of this dependency can be observed in problem C.2, where the prompt directly requires the use of the symbol \(\beta\) defined in B.1 for the calculation of an unknown quantity. This requirement not only tests the participants' ability to understand and apply physical concepts but also assesses their skill in navigating through and linking various parts of a problem set. Such explicit instructions are crucial for guiding participants through the logical progression of the problems, yet the majority of dependencies remain implicit, demanding a deeper level of comprehension and integration of the material.

This structure of problem-solving reflects a realistic scientific inquiry, where discoveries and solutions often rely on previously established knowledge. The explicit mention of \(\beta\) in C.2 as derived from B.1 is emblematic of this educational approach, aiming to foster a holistic understanding and the ability to build upon existing information to solve complex problems. It underscores the importance of thorough comprehension of earlier sections for successful problem-solving in later sections, simulating real-world scientific challenges where new solutions are often predicated on a foundation of established knowledge.

\begin{table*}[ht]
\centering
\begin{tabular}{lcc}
\hline
\toprule
\textbf{Subject} & \textbf{Acronym} & \textbf{Full name}                                \\ \hline
\multirow{5}{*}{Maths} &IMO     & International Mathematical Olympiad      \\
&RMM     & Romanian Master of Mathematics           \\
&ARML    & American Regions Mathematics League      \\
&EMC     & Euclid Mathematics Competition           \\
&EGMO    & European Girls’ Mathematical Olympiad    \\
\hline
\multirow{6}{*}{Physics} &IPhO    & International Physics Olympiad           \\
&APhO    & Asian Physics Olympiad                   \\
&EPhO    & European Physics Olympiad                \\
&USAPhO  & USA Physics Olympiad                     \\
&PUPC    & Princeton University Physics Competition \\ 
&OPhO    & Online Physics Olympiad \\ \hline
\end{tabular}
\caption{Full names of all competitions'acronyms used in this paper}
\label{acronym illustration}
\end{table*}


\section{Evaluation Details}
\subsection{Details of the Evaluated Models}
\subsubsection{LMMs}
We have selected current mainstream LMMs that have performed the best on past scientific multimodal datasets for evaluation.

The closed-source models include: GPT-4V~\cite{2023GPT4VisionSC}, developed by OpenAI, which is currently the most powerful multimodal model. Gemini~\cite{geminiteam2023gemini} is the LMM series developed by Google Deepmind, with Gemini-Ultra-Vision being purported to have surpassed GPT-4V on datasets like MMMU. However the unavailability of Google's API for Gemini Ultra, we test the accessible Gemini-Pro-Vision as an alternative. Qwen-VL-Max~\cite{bai2023qwen}, developed by Alibaba, is the largest LMM, and stands on par with GPT-4V and Gemini-Ultra in multi-modal tasks. Due to the large proportion of Chinese data used in its training, Qwen-VL-Max has a certain advantage in Chinese language ability.

The open-source models include: Yi-VL-34B~\cite{01ai2024Yi} is the first open-source 34B multi-modal model that has demonstrated satisfying performance on several latest datasets. With Chinese text-image pairs included in the training process, Yi-VL-34B offers adequate multilingual support. LLaVA-NeXT-34B~\cite{liu2024llavanext} claims to be the strongest open-source LMM, with enhancements in reasoning, OCR, and world knowledge. Despite being trained exclusively with English multi-modal data, it demonstrates an emergent zero-shot Chinese multi-modal capability on Chinese benchmarks. 

It should be noted that an image must be passed for Gemini-Pro-Vision, LLaVA-NeXT, and Yi-VL during inference. Therefore, for the text-only questions in OlympiadBench dataset, we use the corresponding text-model api (for closed-source models), or their base LLM (for open-source models). To examine the impact of replacing LMM with base LLM for processing text-only data, we subsequently compare the performance differences between GPT-4V and GPT-4 on text-only questions in OlympiadBench.

\subsubsection{LLMs}
The field of LLM starts early in scientific areas such as mathematics and physics, with models specifically trained occurring. We select DeepSeekMath-7B-RL~\cite{shao2024deepseekmath} as the primary baseline for text-only questions. DeepSeekMath-7B-RL is pre-trained on 120B math-related data and enhanced chain-of-thought (CoT) reasoning capabilities using reinforcement learning, in the result scoring close to GPT-4 and Gemini-Ultra on the MATH~\cite{hendrycks2021measuring} dataset. We report the results of the selected LMMs (or their LLM counterparts) on the text-only questions for comparison, and additionally evaluate GPT-4 in order to compare with GPT-4V~\footnote{The version of GPT-4 is "0125-preview" and GPT-4V is "1106-vision-preview".}.




\subsection{Detailed Experiment Result}
\label{subsec:detailed-result}
The comparison of the performance of mainstream closed-ended models on different datasets are shown in Table~\ref{tab:benchmark_comparison2}. 
\begin{table}[htbp]
\centering
\small
\renewcommand{\arraystretch}{1.2}
\begin{tabular}{lccc}
\toprule
                    \multirow{2}{*}{\textbf{Benchmark}} & \multirow{2}{*}{\textbf{GPT-4(V)}} & \multicolumn{1}{c}{\textbf{Qwen}} & \multicolumn{1}{c}{\textbf{Gemini}} \\
                    &  & \textbf{VL-Max} & \textbf{Pro} \\
\hline
MATH                & 52.9     & -         & 32.6       \\
MathVista(testmini) & 49.9     & 50.0         & 45.2       \\
\hline
\textbf{OlympiadBench}       & 17.97        & 10.09            & 4.22          \\
\bottomrule
\end{tabular}
\caption{Comparison of Performance on Different Benchmarks. The values for MATH and MathVista are obtained from Gemini and Qwen's report.}
\label{tab:benchmark_comparison2}
\end{table}

To further discuss the performance difference between GPT-4 and GPT-4V on the Physics-En\_COMP, we split the \textbf{En\_COMP} dataset into two sub-datasets, with \textbf{normal-PhO} being normal PhO questions, and \textbf{long-PhO} being PhO questions that show in a relational series, therefore having long context. As shown in table~\ref{tab:en-comp}, GPT-4 keeps performing slightly better on \textbf{normal-PhO}, but lags much behind on \textbf{long-PhO}, which may indicate improvement of long-context text reasoning capabilities after multimodal training.

\begin{table}[htbp]
    \centering
\begin{tabular}{@{}ccc@{}}
\toprule
\textbf{} & \textbf{\begin{tabular}[c]{@{}c@{}}long-PhO\\ (157)\end{tabular}} & \textbf{\begin{tabular}[c]{@{}c@{}}normal-PhO\\ (74)\end{tabular}} \\ \midrule
GPT-4V    & 18.47                                                             & 1.35                                                               \\
GPT-4     & 14.92                                                             & 4.05                                                               \\ \bottomrule
\end{tabular}
    \caption{Average accuracy of GPT-4V and GPT-4 for the En\_COMP dataset}
    \label{tab:en-comp}
\end{table}

\subsection{Unavailable Responses for Closed-Source Models}
\label{subsec:unavailable-answer}
As described in table~\ref{tab:exp_results}, the response for some problems are not available, the main causes are as follows:
\begin{enumerate}
    \item Exceeding input limit: Some of the context of the problems are too long, which exceed the input token limitation for the API. This case mainly occurs in Physics-En\_COMP that contains long-context problems of over 6,000 tokens.
    \item Inappropriate response: Some problems trigger inappropriate response, which are banned by the API to return.
    \item No response: Some problems continuously get no or empty response from the API.
    \item Request timed out: Some problems continuously fail to get a response.
\end{enumerate}
We removed the problems with unavailable response when calculating the accuracy.

\section{Additional Analysis and Examples}
\label{sec:additional-analysis}

\subsection{Performance analysis of GPT-4V}
\begin{figure*}[ht]
    \centering
    \includegraphics[width=1.0\textwidth]{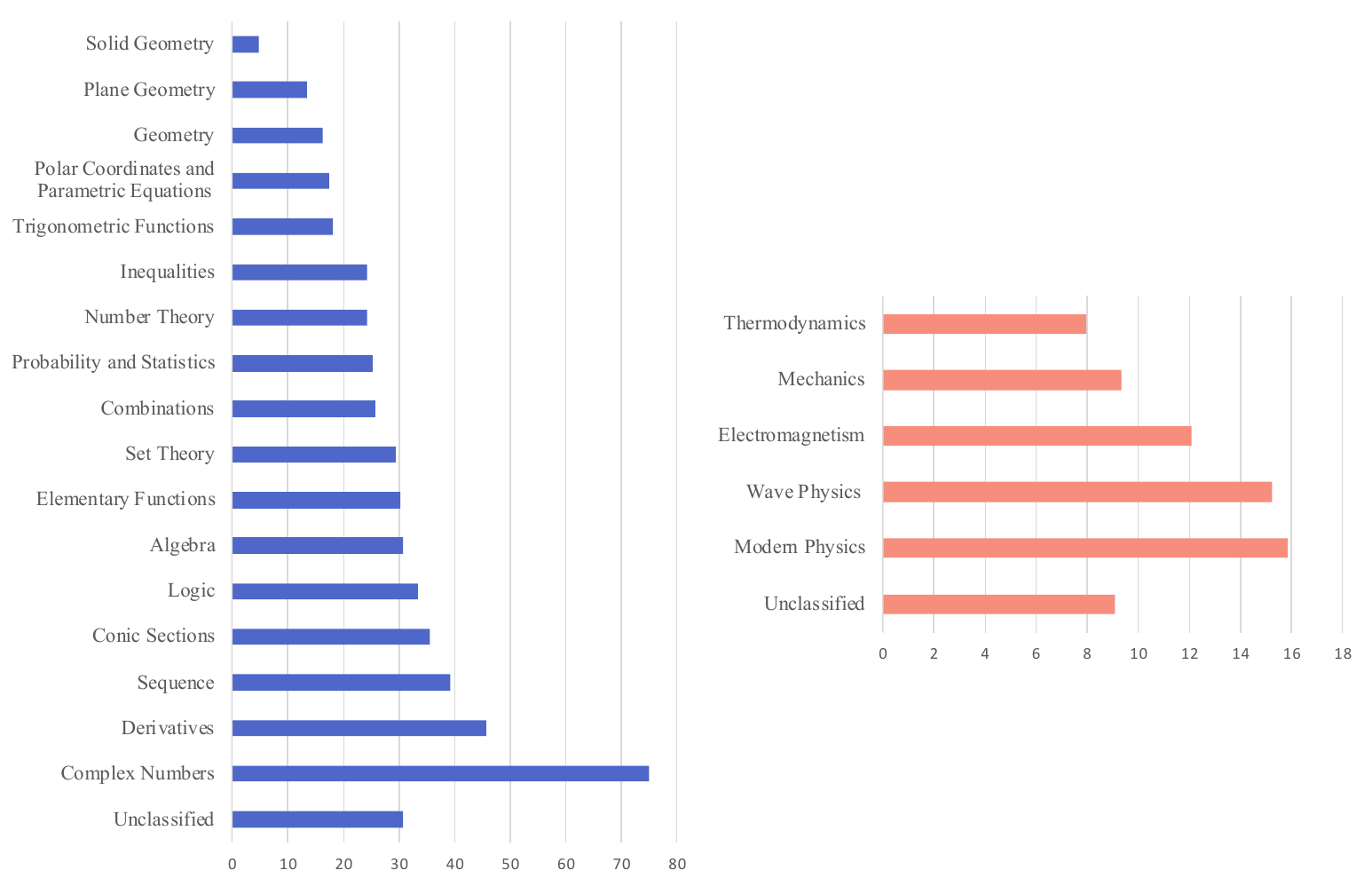}
    \caption{An analysis of GPT-4V's accuracy on different knowledge points, blue for Math and red for Physics.}
    \label{fig:analysis-kp}
\end{figure*}

We analyzed GPT-4V's performance (accuracy on open-ended problems) on different knowledge points based on the knowledge point labels in~\OURS, the results can be found at Figure~\ref{fig:analysis-kp}.

For Math problems, GPT-4V has poor performance in geometry, with the lowest scoring knowledge points being almost exclusively geometry-related. This may show the need of improving the ability of understanding and imaging plane or 3d situations. GPT-4V also performs poorly on knowledge points that are more computationally intensive such as conic curves; and struggles to give a complete and comprehensive classification discussion, therefore prone to making mistakes on combinatorial problems. However, GPT-4V is stronger in knowledge related to derivatives and complex numbers.

As for Physics problems, none of the knowledge points surpass an accuracy of 16\%, and GPT-4V struggles more in thermodynamics and mechanics.

\subsection{Detailed Description of the Error Types in GPT-4V's solving or proving process}
\label{subsec:gpt-4v-error}
\begin{figure*}[ht]
    \centering
    \includegraphics[width=1.0\textwidth]{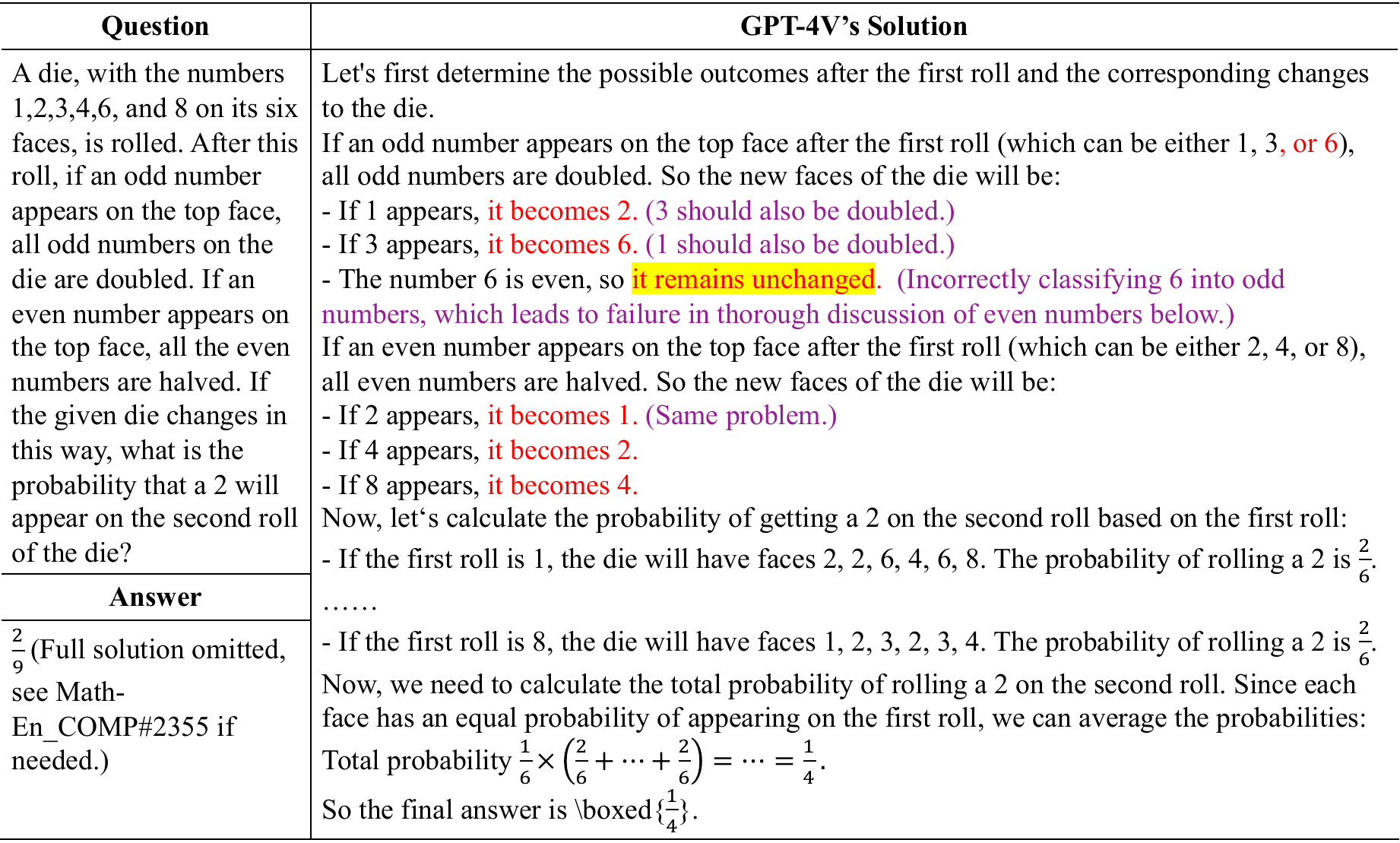}
    \caption{An example of GPT-4V making insufficient classification discussions in a combinatorial problem. The red text is the wrong part of GPT-4V's answer, and the purple text is our further illustration.}
    \label{fig:analysis-classification}
\end{figure*}

\begin{figure*}[ht]
    \centering
    \includegraphics[width=1.0\textwidth]{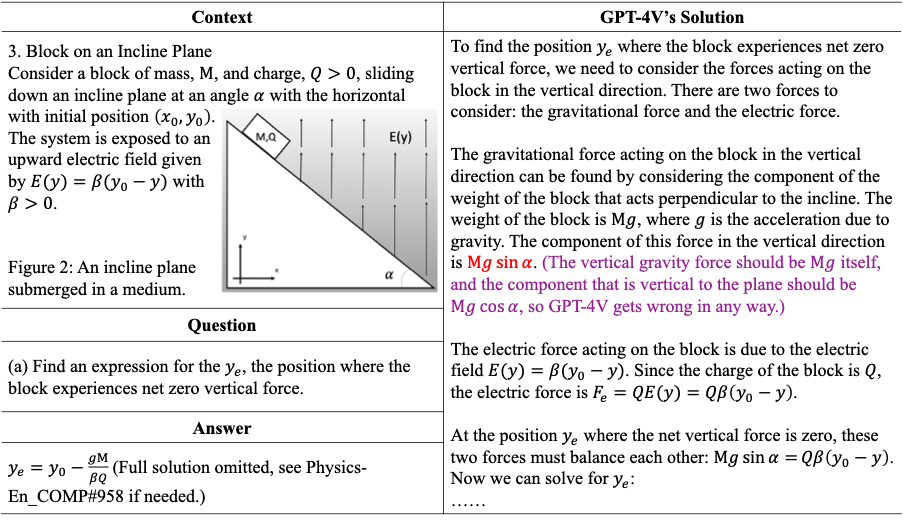}
    \caption{An example of GPT-4V getting confused between the concept of "vertical to the plane" and "vertical(to the ground)".}
    \label{fig:analysis-conceptual}
\end{figure*}

The error types are as follows:
\begin{enumerate}
    \item Question Misunderstanding: GPT-4V sometimes misunderstands the intention or settings of the question.
    \item Value Calculation Error: GPT-4V make simple calculation mistakes sometimes, such as outputting $\frac{b}{2}+7=\frac{b+7}{2}$, these mistakes appears more in Chinese and Math contents.
    \item Expression Calculation Error: Similar to value calculation error, but happens when transforming between two expressions.
    \item Logical Reasoning / Induction Error / Conceptual Confusion: GPT-4V sometimes makes false reasoning or induction, as well as encounters conceptual confusion (see Figure~\ref{fig:analysis-conceptual} for example).
    \item Introducing Unnecessary variables or concepts: GPT-4V sometimes suddenly introduce variables or try to use concepts that have no contribution to solving the problem, which not only makes the output longer, but also may confuse GPT-4V itself and leads to incorrect output.
    \item Conclusion Hallucination: GPT-4V sometimes hallucinates for a conclusion that is not reached in former output, or hallucinates a theorem that does not really exist (for example, when solving geometric proving problem, GPT-4V always mention "The Power Theorem", which does not exist, and all the proof thereafter will lost their logic).
    \item Unfinished Answering: GPT-4V sometimes says the question have confliction in settings (which is not true), or degenerates after some tokens.
    \item Insufficient Classification Discussions: When doing classification discussion, GPT-4V may miss some possible situation, or have overlapped discussion (see Figure~\ref{fig:analysis-classification} for example).
    \item Incorrect Judging: Sometimes GPT-4V gives the right answer, but is judged as incorrect due to the limitation of the automated scoring system: One important problem is that many problems, especially Physics problems, accept answers that fall in a specific range due to rounding up, rather than a fixed numerical answer, so a precision is needed for automatically calculating the range, which was not given in many cases. Using relative precision (such as a percentage of 1\%), may leads to accepting answers with unacceptable error when the origin answer is big, so we manually decide the precision for~\OURS problems if there is no official precision given, which may leads to incorrect judging. For example, in a Physics question (Physics-En\_COMP\#995), the precision is manually set as $1e1$, with the official answer being $\sqrt{2} \times 3970 \approx 5614.43$. GPT-4V calculates the answer as $\sqrt{2} \times 3970 \approx 1.414 \times 3970 \approx 5615.88$, which falls out of the acceptable range, but is actually correct. Another problem is the current automated scoring system can't judge the equality between expressions such as $a\sqrt{b}$ and $\sqrt{a^{2}b}$ with $a \geq 0$. These flaws call for future work such as automatically deciding required precision of the answer, and automatically judging equality between expressions.
\end{enumerate}

\subsection{Other Observation for Open-Ended Problems}
\label{subsec:observations}
\begin{figure*}[ht]
    \centering
    \includegraphics[width=1.0\textwidth]{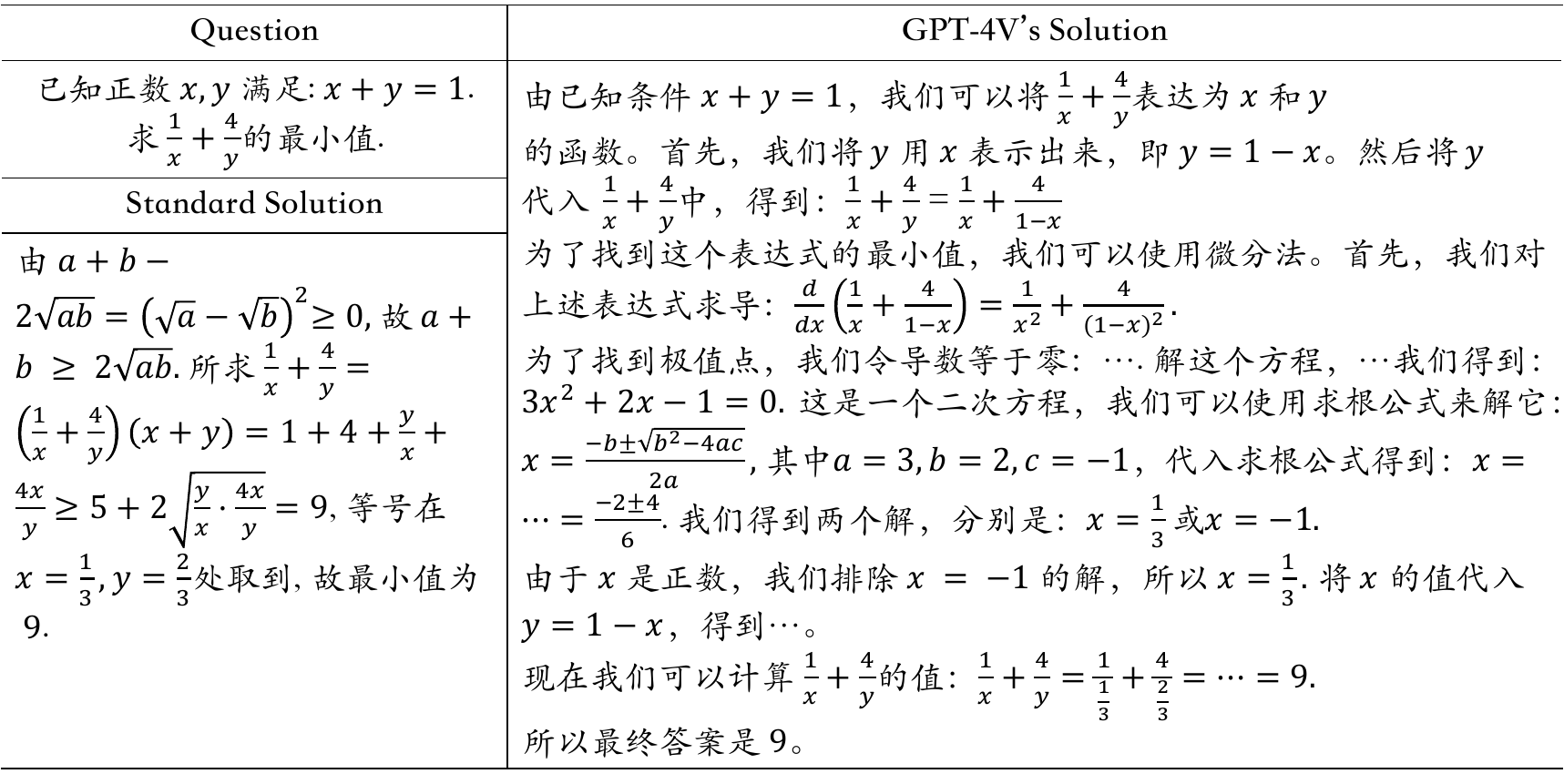}
    \caption{An example of GPT-4V giving a solution that is much complex than that of the standard solution, although reaching the correct answer eventually.}
    \label{fig:analysis-complex}
\end{figure*}

\begin{figure*}[ht]
    \centering
    \includegraphics[width=0.4\textwidth]{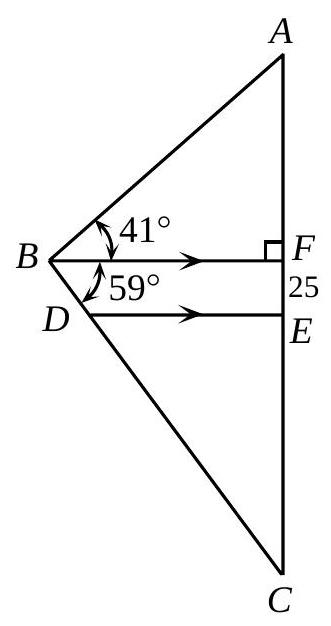}
    \caption{An example of GPT-4V's failure in utilizing image information from Math-Zh\_COMP. GPT-4V starts proving with "we have $\angle DEF = \angle FBC = 59^\circ$", which is an error that can evidently be identified from the image, showing insufficient comprehension of the given plane geometry figure.}
    \label{fig:analysis-geo}
\end{figure*}

\begin{figure*}[ht]
    \centering
    \includegraphics[width=\linewidth]{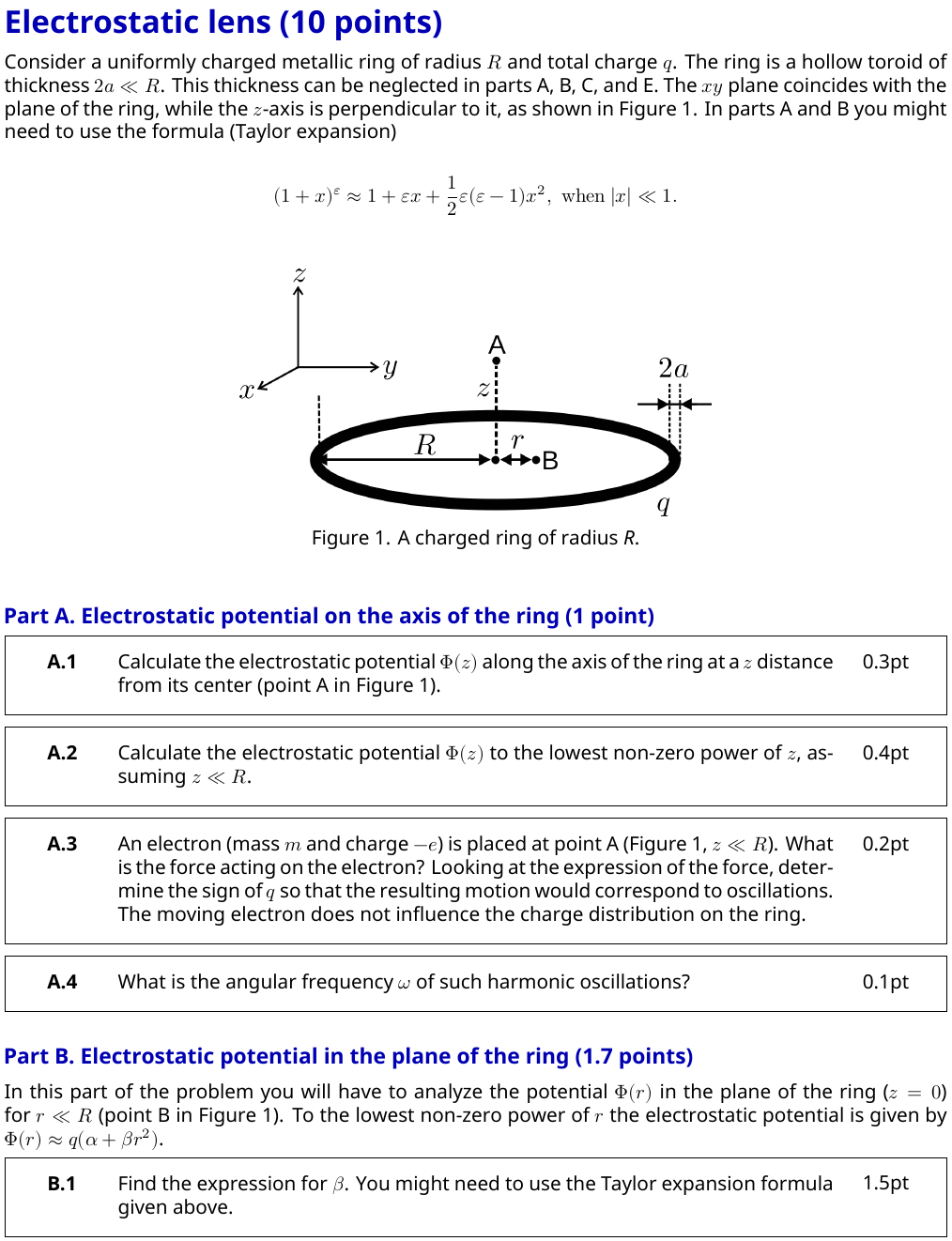}
    \caption{An example illustrating the first section of Problem 2 in IPhO 2021.}
    \label{fig:ipho2021-problem2-part1}
\end{figure*}

\begin{figure*}[ht]
    \centering
    \includegraphics[width=\linewidth]{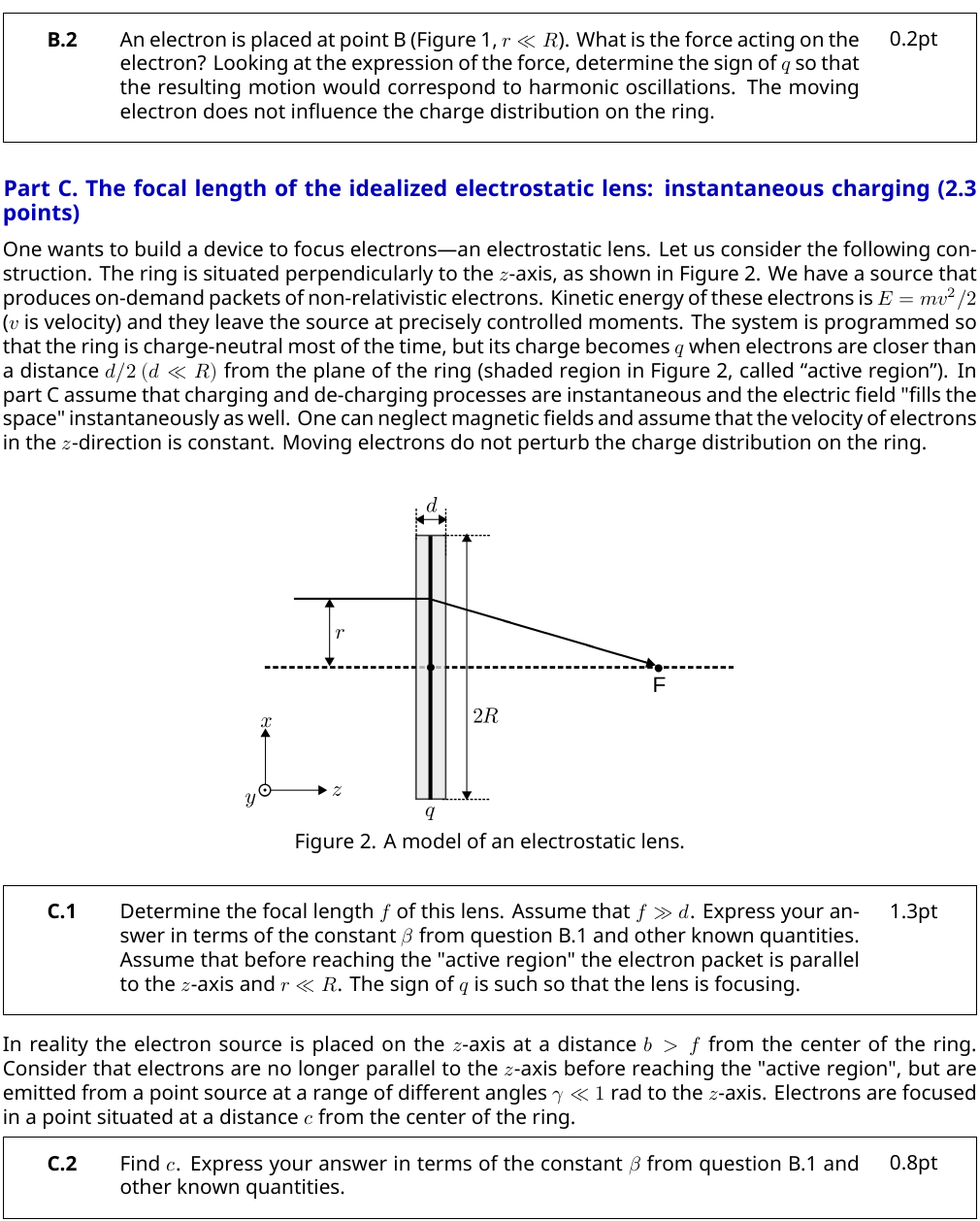}
    \caption{An example illustrating the second section of Problem 2 in IPhO 2021.}
    \label{fig:ipho2021-problem2-part2}
\end{figure*}

\begin{enumerate}
    \item Given a simple solution, GPT-4V may choose a more complex method to solve the problem (see Figure~\ref{fig:analysis-complex}).
    \item Models may give correct answers with a false process. Mainly observed for problems with a simple answer, such as the variables takes $0$ as the answer.
    \item GPT-4V may success in giving correct overall idea, but fail in calculation (such as solving quadratic equations with extra negative signs), which leads to a wrong answer.
    \item GPT-4V may not fully utilize the information from the image (see Figure~\ref{fig:analysis-geo}).
\end{enumerate}

\begin{table*}[ht]
\centering
\small
\setlength{\tabcolsep}{2.3pt}
\renewcommand{\arraystretch}{1.3}
\begin{tabular}{lcc}
\toprule
\textbf{Subset} & \textbf{Subfield} & \textbf{Number} \\ \midrule
\multirow{13}{*}{\textbf{CEE Math}}      
             & Derivative             & 334    \\
             & Conic Sections         & 350    \\
             & Sequence               & 258    \\
             & Trigonometric Functions & 236   \\
             & Set Theory             & 22     \\
             & Logic                  & 3      \\
             & Elementary Functions   & 158    \\
             & Inequality             & 138    \\
             & PC\&PE                 & 80     \\
             & Probability and Statistics & 758 \\
             & Plane Geometry         & 790    \\
             & Solid Geometry         & 1322   \\
             & Complex Numbers        & 8      \\
\hline
\multirow{4}{*}{\textbf{COMP Math}}     
             & Combinatorics          & 369    \\
             & Algebra                & 525    \\
             & Number Theory          & 256    \\
             & Geometry               & 535    \\
\hline
\multirow{5}{*}{\textbf{CEE\&COMP Physics}} 
             & Mechanics              & 1010   \\
             & Electromagnetism       & 714    \\
             & Thermodynamics         & 248    \\
             & Optics                 & 153    \\
             & Modern Physics         & 209    \\ \bottomrule
\end{tabular}
\caption{Statistics of subfield in Mathematics and Physics. PC\&PE stands for Polar Coordinates and Parametric Equations.}
\label{tab:number-subfield}
\end{table*}

\section{Automatic Scoring Pipeline}
The pipeline workflow is shown in Algorithm~\ref{pipeline algorithm}.

\begin{algorithm*}
\caption{Auto Scoring Judge}
\SetAlgoLined
\textbf{Input:} GroundTruth, ModelOutput\;
\textbf{Output:} Boolean value indicating match\;

\BlankLine
Preprocess GroundTruth and ModelOutput\;

\eIf{GroundTruth equals ModelOutput}{
    \Return True\;
}{
    \eIf{GroundTruth and ModelOutput are intervals or tuples}{
        \eIf{GroundTruth equals ModelOutput as intervals or tuples}{
            \Return True\;
        }{
            \Return False\;
        }
    }{
        \eIf{GroundTruth equals ModelOutput numerically}{
            \Return True\;
        }{
            \eIf{Both GroundTruth and ModelOutput contain "="}{
                \Return EquationEqual(GroundTruth, ModelOutput)\;
            }{
                \Return ExpressionEqual(GroundTruth, ModelOutput)\;
            }
        }
    }
}

\label{pipeline algorithm}
\end{algorithm*}

\end{document}

%% file: experiment.tex
\section{Experiments}
\subsection{Settings}

\begin{figure*}[ht]
    \centering
    \includegraphics[width=1\linewidth]{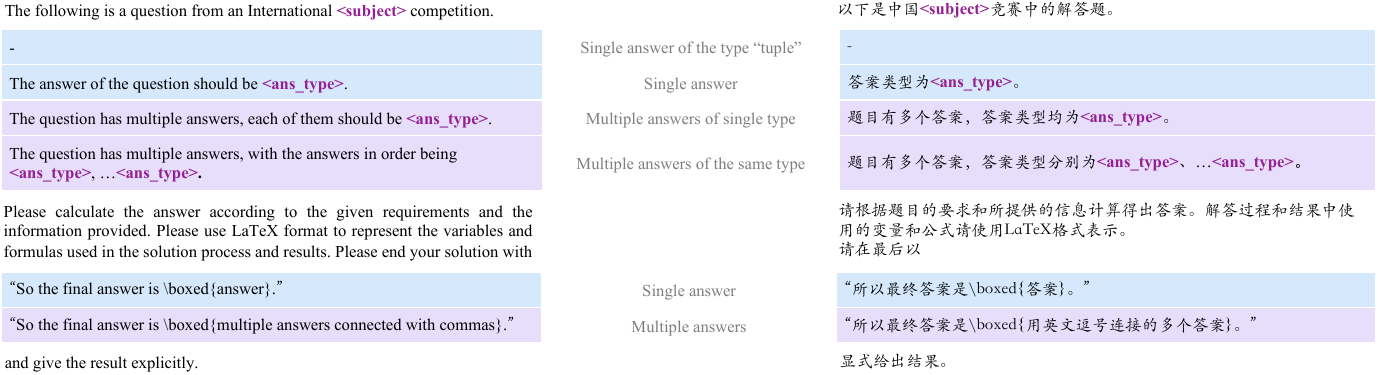}
    \caption{The template of the construction of the prompt for English(left) and Chinese(right) open-ended questions, among which <subject>, <ans\_type>, and whether there are multiple answers can all be obtained from the data items in OlympiadBench dataset.}
    \label{fig:en-prompt}
\end{figure*}

We conduct evaluations of open-source and closed-sourced LMMs that have been selected with consideration of their comprehensive capabilities on OlympiadBench. At the same time, we have selected LLMs with strong mathematical and logical abilities for evaluation on text-only questions.

As no accurate automatic evaluation method for theorem proving exists, we run full experiment on the automatic-scoring-available open-ended problems with answer type included in the Table \ref{examples of answer types}, which is discussed in this section. We do manual sampling check of GPT-4V for theorem proving problems with analysis reported at Section~\ref{subsec:proving}.
\subsubsection{Prompts}
We evaluate the models in a zero-shot setting. Due to the high difficulty of the OlympiadBench questions, there should be considerable randomness in the results when using small batch data as the validation set, so we directly use a specific prompt template for all models instead of conducting prompt-engineering for each model respectively. 
The prompt template for English and Chinese open-ended questions is shown in the figure~\ref{fig:en-prompt}. 
To ensure the most complete extraction of the model's final results, we explicitly prescribe the types and formats of the answers in the prompt to promote the accuracy of the machine's automatic scoring.
In order to test the native mathematical and physical abilities of the models, the prompts used in the test do not introduce knowledge points and other extra information contained in the dataset, but this information can be applied in subsequent research. Note that deepseek-math-7B-RL~\cite{shao2024deepseekmath} requires the addition of a specific chain-of-thought prompt at the end of the input, which we adhered to during the evaluation.


\subsubsection{Evaluation Workflow}
We first apply each model to generate answers for questions in OlympiadBench using prompts formed by prompt template, with open-source models running on NVIDIA A800 GPUs. Then, we run the automatic scoring pipeline to judge the correctness of the answers as described in subsection~\ref{subsec:pipeline}. Finally, we calculate the micro-average accuracy as the comparing metric.
The code of the whole workflow is provided in the supplementary material.

\subsection{Baselines}

In our study, we evaluate the performance of current leading bilingual large multimodal models (LMMs), as well as bilingual large language models (LLMs) that has strong mathematical and reasoning abilities. We take both open- and closed-source models into consideration, using either the largest and latest released checkpoints or the best-performing official APIs available.

For LMMs, we select GPT-4V(GPT-4-Vision)~\cite{2023GPT4VisionSC}, Gemini-Pro-Vision~\cite{geminiteam2023gemini}, Qwen-VL-Max~\cite{bai2023qwen} for closed-source models, while Yi-VL-34B~\cite{01ai2024Yi} and LLaVA-NeXT-34B~\cite{liu2024llavanext} for open-source models. For models that demand compulsory image input, we take their LMM counterpart (corresponding text-model API or base LLM) for evaluation. 
Specifically, for LLaVA-NeXT-34B, we use its base LLM, Nous-Hermes-2-Yi-34B~\cite{Nous2023Nous}. The text model corresponding to Yi-VL-34B is Yi-34B-Chat~\cite{01ai2023Yi}. Similarly, for the Gemini-Pro-Vision, we utilize the Gemini-Pro API interface.
To examine the impact of replacing LMM with base LLM for processing text-only data, we subsequently compare the performance differences between GPT-4V and GPT-4~\footnote{The version of GPT-4 is "0125-preview" and GPT-4V is "1106-vision-preview".} on text-only questions in OlympiadBench.

For LLMs, we select DeepSeekMath-7B-RL~\cite{shao2024deepseekmath} as the primary baseline for text-only questions, and report the results of the selected LMMs (or their LLM counterparts) on the text-only questions for comparison, and additionally evaluate GPT-4 as described above.

\subsection{Main Results}

\begin{table*}[ht]
\centering
\small
\renewcommand{\arraystretch}{1.2}
\begin{tabular}{lccccccccc}
\toprule
\multirow{2}{*}{\textbf{Models}} & \multicolumn{4}{c}{\textbf{Maths}} & \multirow{2}{*}{} &  \multicolumn{3}{c}{\textbf{Physics}} & \multirow{2}{*}{\textbf{Avg.}}\\
\cline{2-5} \cline{7-9}
\multicolumn{1}{c}{}                        & \textbf{En\_COMP} & \textbf{Zh\_COMP} & \textbf{Zh\_CEE} & \textbf{Avg.} &  & \textbf{En\_COMP}      & \textbf{Zh\_CEE}    & \textbf{Avg.} &  \\
\hline
LLaVA-NeXT-34B\dag 
& 3.98     &  2.60  &    4.64     & 4.30  & - & 1.36     &   2.32        &  2.08 & 3.65  \\
Yi-VL-34B\dag 
& 4.22 & 3.68 & 4.31 & 4.23 & - & 0.91 & 1.64 & 1.46  & 3.42  \\
\hdashline
Gemini-Pro-Vision 
& 6.92     & 2.59     & 5.05*   & 5.14                 & - & 3.19*    & 2.12    & 2.45                 & 4.22   \\
Qwen-VL-Max         & 10.68   & 13.21*    & 13.08  & 12.65                & - & 3.76*    & 5.64*   & 5.09                 & 10.09   \\
GPT-4V        & 27.18   & 14.87    & 21.27   & 21.70                & - & 11.42    & 10.45   & 10.74                & 17.97   \\
\hline
\multicolumn{10}{c}{Experiment with text-only} \\ \hline
LLaVA-NeXT-34B          & 4.15     &   2.94       &    8.55     &  6.29    & - & 2.12     & 5.22 &   3.13    & 5.87  \\
Yi-VL-34B               & 4.45     & 3.68      & 8.06    & 6.24  & - & 0.85     & 5.22      & 2.28  &  5.72 \\ 
DeepSeekMath-7B-RL            & 19.44    & 2.70     & 22.42   & 18.09 & - & 6.78     & 16.52     & 9.97  & 17.02   \\
\hdashline
Gemini-Pro-Vision          & 7.57     & 2.94     & 9.20*  & 7.63  & - & 4.66     & 6.96      & 5.41  & 7.34  \\
Qwen-VL-Max         & 11.57   & 14.29    & 25.89   & 19.70 & - & 4.24    & 18.26     & 8.83 & 18.27   \\
GPT-4V              & 28.93   & 15.93    & 37.10  & 31.01 & - & 12.71    & 23.48     & 16.24 & 29.07   \\
GPT-4               & 30.42   & 16.42     & 37.98   & 32.00 & - & 12.29    & 24.35     & 16.24 & 29.93   \\
\bottomrule
\end{tabular}
\caption{Experimental results. En\_COMP: COMP problems in English, Zh\_COMP: COMP problems in Chinese, Zh\_CEE: CEE problems in Chinese. For closed-source models, the responses for some problems are not available, we mark the results with * (all of the proportion of missing answers are less than 1\%, except for the result of Qwen-VL-Max in Physics-En\_COMP, where  26 questions exceed maximum input length). The causes are further described in Appendix~\ref{subsec:unavailable-answer}. Moreover, LLaVA-NeXT-34B and Yi-VL-34B only accepts input with single image, we mark results from only one image input with \dag.}
\label{tab:exp_results}
\end{table*}


The overall experiment result is shown in table~\ref{tab:exp_results}. Based on the results, our key findings can be summarized as the following:

\textbf{OlympiadBench is more challenging than existing benchmarks, which provides new perspective to compare LMMs.} As shown in table~\ref{tab:benchmark_comparison2}, the most advanced model only achieves an average accuracy of 17.97\% on OlympiadBench, which is much lower than that of existing benchmarks. Moreover, the gap between the models has been widened, thereby becoming more significant, which helps people to compare the differences in capabilities between different models more accurately.

\textbf{There still exists a huge difference between the most powerful closed-source models and open-source models, but a large model size is needed.} The average accuracy of GPT-4V is more than 5 times larger than the best-performing open-source model (Yi-VL-34B). But Gemini-Pro-Vision, being closed-source models of the second-tier size, is much less compatible on complicated tasks such as OlympiadBench, for it achieves an average accuracy that is only slightly higher than open-source model.

\textbf{The challenge lies more on question-with-images, Physics and none-English text.} The model performance on text-only questions is significantly above average, showing the challenging spirit of multi-modal questions. Meanwhile, Physics questions, especially Physics questions with images, are more challenging than math questions, as they require knowledge of the laws of Physics as well as other world knowledge besides mathematical abilities such as calculation and reasoning. 
Moreover, LMMs with a focus on bilingual image-text training data, such as Qwen-VL-Max and Yi-VL-34B, perform better on Chinese questions then English questions.

\textbf{Open source LLMs is catching at fast speed in the area of maths and physics.} Although with a relatively small size, DeepSeekMath-7B-RL outperforms or is on par with Gemini-Pro-Vision and Qwen-VL-Max on the text-only part of OlympiadBench, especially in Math problems, showing promising future of open-source model of pre-training and fine-tuning on fine-grained mathematical and reasoning data.

\textbf{Multi-modal training slightly hurts performance on text-only math and physics tasks, but may also bring some improvement.} The text-only version GPT-4 performs slightly better on all datasets of OlympiadBench, except for the \textbf{En\_COMP} dataset. We hypothesis that the improvement in the \textbf{En\_COMP} dataset shows an enhancement of long-context text reasoning capabilities, which is discussed in Appendix~\ref{subsec:detailed-result}.

%% file: analysis.tex
\section{Analysis}
In this section, we conduct analysis on the GPT-4V's answers of specific open-ended questions that have been sampled, as well as giving preliminary examination of theorem proving questions.

\subsection{Examination of Theorem Proving Questions}
\label{subsec:proving}
For GPT-4V, we do manual sampling check to evaluate the mathematical theorem proving questions. In the questions drawn according to the knowledge point distribution, GPT-4V only answers 6 out of 81 questions correctly in Math-Zh\_COMP, all of which are relatively simple and classic conclusions (e.g. AM-GM inequality), or involved only simple computational derivations, and was basically unable to complete the proof within the token limitation in Math-En\_COMP, indicating that existing models still cannot effectively solve lengthy reasoning and proofs, which is consistent with the conclusions in existing papers~\cite{trinh2024solving}.

In solving proof problems, GPT-4V exposes several important issues, including: inability to fully utilize image information (figure~\ref{fig:analysis-geo} as an example); tending to make mistakes in simplifying and transforming algebraic expressions; proposing simple, basic incorrect conclusions;struggling with classification discussions, etc. Detailed examples can be found in the Appendix~\ref{sec:additional-analysis}.

\subsection{Mistake Analysis of GPT-4V}
\begin{figure}[ht]
    \centering
    \includegraphics[width=1.0\linewidth]{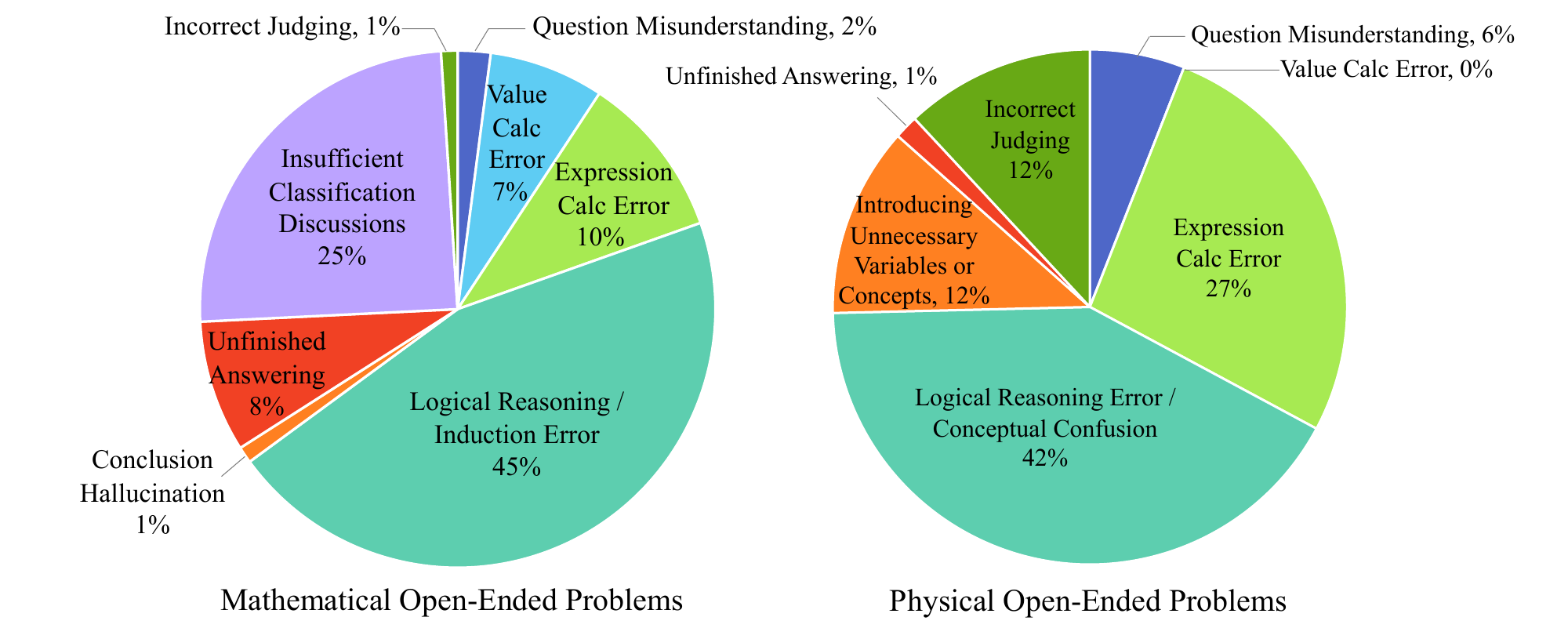}
    \caption{Distribution of the error occurring in GPT-4V's solving process of 164 sampled Olympic-level open-ended problems.}
    \label{fig:analysis-open-ended-error}
\end{figure}
We manually sample and check 97 maths (55 for English and 42 for Chinese) and 67 physics Olympiad-level open-ended problems that GPT-4V fails, and analyze the type of mistakes, the overall results are shown in figure~\ref{fig:analysis-open-ended-error}.
In maths problems, the typical errors of GPT-4V include: insufficient classification discussion, especially in combinatorial problems; poor performance in problems requiring large calculations (e.g. conic curve problems), manifests as a lack of logic in the calculation process, resulting in the model being unable to provide a reasonable answer. However, we also found that GPT-4V has strong abilities in solving quadratic equations and derivative problems. 
In physics problems, GPT-4V tends to fall in conceptual confusion, or introduce unnecessary variables or concepts, but its capability to simplify and transform algebraic expressions is stronger than in purely mathematical situations, with nearly no numerical calculation errors.
